\documentclass[11pt,a4paper]{article}
\usepackage{times,latexsym}
\usepackage{url}
\usepackage[T1]{fontenc}

\usepackage[]{style/acl2019}
\usepackage{style/mystyle}
\usepackage{style/symbol}
\usepackage{style/mlsymbols}
\usepackage{hyperref}
\usepackage{float}

\newtheorem{mydef}{Definition}

\aclfinalcopy 

\newcommand{\theTitle}{Augmenting Neural Networks with First-order Logic}

\title{\theTitle}

\author{Tao Li \\
  University of Utah \\
  {\tt tli@cs.utah.edu} \\\And
  Vivek Srikumar \\
  University of Utah \\
  {\tt svivek@cs.utah.edu} \\}

\date{}

\begin{document}
\maketitle

\begin{abstract}
  Today, the dominant paradigm for training neural networks involves
  minimizing task loss on a large dataset.
Using world knowledge to inform a model, and yet retain the ability to perform end-to-end training remains an open question.
In this paper, we present a novel framework for introducing
declarative knowledge to neural network architectures
in order to guide training and prediction.
Our framework systematically compiles logical statements into computation graphs that
augment a neural network without extra learnable parameters or manual redesign.
We evaluate our modeling strategy on three tasks: machine
comprehension, natural language inference, and text chunking.
Our experiments show that knowledge-augmented networks can strongly improve over baselines, especially in low-data regimes.


\end{abstract}


\section{Introduction}
\label{sec:intro}

Neural models demonstrate remarkable predictive performance across a
broad spectrum of NLP tasks: \eg, natural language
inference~\cite{parikh2016decomposable}, machine
comprehension~\cite{seo2016bidirectional}, machine
translation~\cite{bahdanau2014neural}, and
summarization~\cite{rush2015neural}. These successes can be
attributed to their ability to learn robust representations from
data. However, such end-to-end training demands a large number of
training examples; for example, training a typical network for
machine translation may require millions of sentence
pairs~\cite[\eg][]{luong2015effective}.  The difficulties and
expense of curating large amounts of annotated data are well
understood and, consequently, massive datasets may not be available
for new tasks, domains or languages.

\begin{figure}
  \centering
  \fbox{\includegraphics[width=0.92\linewidth]{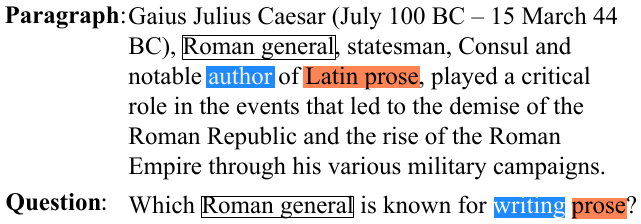}}
  \caption{An example of reading comprehension that
    illustrates alignments/attention. In this paper, we consider the
  problem of incorporating external knowledge about such alignments
  into training neural networks.}
  \label{fig:example}
\end{figure}

In this paper, we argue that we can combat the data hungriness of
neural networks by taking advantage of domain knowledge expressed as
first-order logic. As an example, consider the task of reading
comprehension, where the goal is to answer a question based on a
paragraph of text (Fig.~\ref{fig:example}). Attention-driven
models such as BiDAF~\cite{seo2016bidirectional} learn to align
words in the question with words in the text as an intermediate step
towards identifying the answer.  While alignments
(e.g. \example{author} to \example{writing}) can be learned from
data, we argue that models can reduce their data dependence if they
were guided by easily stated rules such as: \example{Prefer aligning
  phrases that are marked as similar according to an external
  resource, \eg, ConceptNet~\cite{liu2004conceptnet}}. If such
declaratively stated rules can be incorporated into training neural
networks, then they can provide the inductive bias that can reduce
data dependence for training.

That general neural networks can represent such Boolean functions is
known and has been studied both from the theoretical and empirical
perspectives~\cite[\eg][]{maass1994comparison,anthony2003boolean,PanSr2016}. Recently,
\citet{hu2016harnessing} exploit this property to train a neural
network to mimic a teacher network that uses structured rules.  In
this paper, we seek to directly incorporate such structured
knowledge into a neural network architecture without substantial
changes to the training methods. We focus on three questions:
\begin{enumerate}
\item Can we integrate declarative rules with end-to-end neural
  network training?
\item Can such rules help ease the need for data?
\item How does incorporating domain expertise compare against
  large training resources powered by pre-trained representations?
\end{enumerate}

The first question poses the key technical challenge we address in
this paper. On one hand, we wish to guide training and prediction
with neural networks using logic, which is non-differentiable. On
the other hand, we seek to retain the advantages of gradient-based
learning without having to redesign the training scheme. To this
end, we propose a framework that allows us to systematically augment
an {\em existing} network architecture using constraints about its nodes
by deterministically converting rules into differentiable
computation graphs. To allow for the possibility of such rules being
incorrect, our framework is designed to admit soft constraints from
the ground up.
Our framework is compatible with off-the-shelf neural networks
without extensive redesign or any additional trainable parameters.



To address the second and the third questions, we empirically
evaluate our framework on three tasks: machine comprehension,
natural language inference, and text chunking. In each case, we use
a general off-the-shelf model for the task, and study the impact of
simple logical constraints on observed neurons (e.g., attention) for
different data sizes. We show that our framework can successfully
improve an existing neural design, especially when the number of
training examples is limited.

In summary, our contributions are:
\begin{enumerate}
\item We introduce a new framework for incorporating first-order
  logic rules into neural network design in order to guide both
  training and prediction.
  
\item We evaluate our approach on three different NLP tasks: machine
  comprehension, textual entailment, and text chunking. We show that
  augmented models lead to large performance gains in the low
  training data regimes.\footnote{The code used for our experiments
    is archived here: \url{https://github.com/utahnlp/layer_augmentation}}
\end{enumerate}



\section{Problem Setup}
\label{sec:prelim}

In this section, we will introduce the notation and assumptions that
form the basis of our formalism for constraining neural networks.

Neural networks are directed acyclic computation graphs $G=(V,E)$,
consisting of nodes (i.e. neurons) $V$ and weighted directed edges
$E$ that represent information flow.
%
Although not all neurons have explicitly grounded meanings,
some nodes indeed can be endowed with semantics tied to the task.
%
%
Node semantics may be assigned during model design (\eg attention),
or incidentally discovered in post hoc analysis~\cite[\eg,][and others]{le2011building,radford2017learning}.
In either case, our goal is to augment a neural network with such \concept{named neurons} using
declarative rules.



The use of logic to represent domain knowledge has a rich history in
AI~\cite[\eg][]{russell2016artificial}. 
%
In this work, to capture such knowledge, we will primarily focus on
conditional statements of the form $L \rightarrow R$, where the
expression $L$ is
the antecedent (or the left-hand side) that can be conjunctions or
disjunctions of literals, and $R$ is the consequent (or the right-hand
side) that consists of a single literal.  Note that such rules
include Horn clauses
and their generalizations, which are well studied in the knowledge
representation and logic programming communities~\cite[\eg][]{chandra1985horn}.




Integrating rules with neural networks presents three
difficulties. First, we need a mapping between the predicates in the rules and nodes in the computation graph. 
Second, logic is not differentiable;
we need an encoding of logic that admits training using gradient based methods.
Finally, computation graphs are
acyclic, but user-defined rules may
introduce cyclic dependencies between the nodes. Let us look at
these issues in order.

As mentioned before, we will assume named neurons are given.
And by associating predicates with such nodes that are endowed with symbolic meaning,
we can introduce domain knowledge about a problem in terms of
these predicates.
In the rest of the paper, we will use
lower cased letters (\eg, $a_i, b_j$) to denote nodes in
a computation graph, and upper cased letters (\eg, $A_i, B_j$) for
predicates associated with them.


To deal with the non-differentiablity of logic, we will treat the
post-activation value of a named neuron as the degree to which the
associated predicate is true. In \S\ref{sec:framework}, we will look
at methods for compiling conditional statements into differentiable
statements that augment a given network.

\paragraph{Cyclicity of Constraints}

Since we will augment computation graphs with compiled conditional forms,
we should be careful to avoid creating cycles.  To formalize this,
let us define cyclicity of conditional statements with respect to a
neural network.

Given two nodes $a$ and $b$ in a computation graph, we say that the
node $a$ is \emph{upstream} of node $b$ if there is a directed path from $a$
to $b$ in the graph.

\begin{mydef}[Cyclic and Acyclic
  Implications]
  \label{def:cyclic-implications}
  
  Let $G$ be a computation graph. An implicative statement
  $L \rightarrow R$ is \emph{cyclic} with respect to $G$ if, for any
  literal $R_i \in R$, the node $r_i$ associated with it is upstream
  of the node $l_j$ associated with some literal $L_j \in L$.
  An implicative statement is \emph{acyclic} if it is not cyclic.
  
\end{mydef}

\begin{figure}
  \centering
  \includegraphics[width=\linewidth]{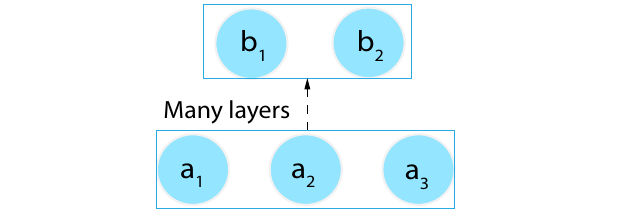}
  \caption{An example computation graph.
  The statement $A_1 \land B_1 \rightarrow A_2 \land B_2$ is cyclic
  with respect to the graph.  On the other hand, the statement
  $A_1 \land A_2 \rightarrow B_1 \land B_2$ is acyclic.}
  \label{fig:flow1}
\end{figure}


Fig.~\ref{fig:flow1} and its caption gives examples of cyclic and
acyclic implications.
A cyclic statement sometimes can be converted to an equivalent acyclic statement
by constructing its contrapositive. For example, the constraint
$B_1 \rightarrow A_1$ is equivalent to
$\neg A_1 \rightarrow \neg B_1$. While the former is cyclic, the later is acyclic.
Generally, we can assume that we have acyclic
implications.\footnote{As we will see in \S\ref{sec:combine}, the
  contrapositive does not always help because we may end up with a
  complex right hand side that we can not yet compile into the
  computation graph.}



\section{A Framework for Augmenting Neural Networks with Constraints}
\label{sec:framework}

To create constraint-aware neural networks, we will extend the
computation graph of an existing network with additional edges
defined by constraints.
In \S\ref{sec:workflow}, we will focus on the case
where the antecedent is conjunctive/disjunctive and the consequent is a single literal.
In \S\ref{sec:general-framework}, we will cover more general antecedents.

\subsection{Constraints Beget Distance Functions}
\label{sec:workflow}
Given a computation graph, suppose we have a acyclic conditional
statement: $Z\rightarrow Y$, where $Z$ is a conjunction
or a disjunction of literals and $Y$ is a single literal.
We define the neuron associated with $Y$ to be $y = g\p{\bW\bx}$,
where $g$ denotes an activation function, $\bW$ are network parameters,
$\bx$ is the immediate input to $y$.
Further, let the vector $\bz$ represent the neurons associated with the predicates in $Z$.
While the nodes $\bz$ need to be named neurons, the
immediate input $\bx$ need not necessarily
have symbolic meaning.

\paragraph{Constrained Neural Layers} Our goal is to augment the
computation of $y$ so that whenever $Z$ is true, the pre-activated
value of $y$ increases if the literal $Y$ is not negated (and
decreases if it is). To do so, we define a {\em constrained neural
  layer} as
\begin{equation}
  \label{eq:constrained-neural-layer}
  y = g\p{\bW\bx +\rho d\p{\bz}}.
\end{equation}
Here, we will refer to the function $d$ as the \emph{distance
  function} that captures, in a differentiable way, whether the
antecedent of the implication holds. The importance of the entire
constraint is decided by a real-valued hyper-parameter $\rho \geq 0$.


The definition of the constrained neural layer says that, by
compiling an implicative statement into a distance function, we can
regulate the pre-activation scores of the downstream neurons based
on the states of upstream ones.

\paragraph{Designing the distance function}
The key consideration in the compilation step is the choice of an
appropriate distance function for logical statements. The ideal
distance function we seek is the indicator for the statement $Z$:
\begin{equation*}
  d_{ideal}(\bz) = \begin{cases}
    1, & \mbox{if $Z$ holds}, \\
    0, & \mbox{otherwise}.
    \end{cases}
\end{equation*}
However, since the function $d_{ideal}$ is not differentiable, we
need smooth surrogates.


In the rest of this paper, we will define and use distance functions
that are inspired by probabilistic soft
logic~\cite[c.f.][]{klement2013triangular} and its use of the
\L{}ukasiewicz T-norm and T-conorm to define a soft version of
conjunctions and disjunctions.\footnote{The definitions of the
  distance functions here as surrogates for the non-differentiable
  $d_{ideal}$ is reminiscent of the use of hinge loss as a surrogate
  for the zero-one loss. In both cases, other surrogates are
  possible.}

Table~\ref{tab:tab1} summarizes distance functions corresponding to
conjunctions and disjunctions. In all cases, recall that the $z_i$'s
are the states of neurons and are assumed to be in the range
$[0,1]$. Examining the table, we see that with a conjunctive
antecedent (first row), the distance becomes zero if even one of the
conjuncts is false. For a disjunctive antecedent (second row), the
distance becomes zero only when all the disjuncts are false;
otherwise, it increases as the disjuncts become more likely to be
true.

\begin{table}
  \centering
  \begin{tabular}{rl}
    \toprule
    Antecedent    & Distance $d(\bz)$                        					\\
    \midrule
    $\Land_i Z_i$      & $ \max(0, \sum_i z_i - |Z| + 1)$                       \\
    $\Lor_i Z_i$       & $ \min(1, \sum_i z_i)$                               \\
    $\neg \Lor_i Z_i$  & $ \max(0, 1 - \sum_i z_i)$                           \\
    $\neg \Land_i Z_i$ & $ \min(1, N - \sum_i z_i)$                           \\
    \bottomrule
  \end{tabular}
  \caption{Distance functions designed using the \L{}ukasiewicz T-norm.
    Here, $|Z|$ is the number
    of antecedent literals.  $z_i$'s are upstream neurons
    associated with literals $Z_i$'s.}
  \label{tab:tab1}
\end{table}

\paragraph{Negating Predicates} 
Both the antecedent (the $Z$'s) and the consequent ($Y$) could
contain negated predicates. We will consider these separately.

For any negated antecedent predicate, we modify the distance
function by substituting the corresponding $z_i$ with $1-z_i$ in Table~\ref{tab:tab1}.
The last two rows of the
table list out two special cases, where the
entire antecedents are negated, and can be derived from the first
two rows. 

To negate consequent $Y$, we need to reduce the pre-activation score of neuron $y$.
To achieve this, we can simply negate the entire distance function.

\paragraph{Scaling factor $\rho$}
In Eq.~\ref{eq:constrained-neural-layer}, the distance function serves to
promote or inhibit the value of downstream neuron. The extent is controlled
by the scaling factor $\rho$.
For instance, with $\rho = +\infty$, the pre-activation score of the
downstream neuron is dominated by the distance function. 
In this case, we
have a hard constraint.  In contrast, with a small $\rho$, the
output state depends on both the $\bW\bx$ and the distance
function. In this case, the \emph{soft} constraint serves more as a suggestion.
Ultimately, the network parameters might overrule the constraint.
We will see an example in \S\ref{sec:experiments} where noisy constraint prefers small $\rho$.

\subsection{General Boolean Antecedents}
\label{sec:general-framework}
So far, we exclusively focused on conditional statements with either
conjunctive or disjunctive antecedents.  In this section, we
will consider general antecedents.

As an illustrative example, suppose we have an antecedent
$(\neg A \vee B) \wedge (C \vee D)$. By introducing auxiliary
variables, we can convert it into the conjunctive form $P \wedge Q$,
where $(\neg A \vee B) \leftrightarrow P$ and
$(C \vee D) \leftrightarrow Q$.  To perform such operation, we need
to:
\begin{inparaenum}[(1)]
\item introduce auxiliary neurons associated with the auxiliary predicates $P$
  and $Q$, and,
\item define these neurons to be exclusively determined by the
  biconditional constraint.
\end{inparaenum}


%
%
To be consistent in terminology, when considering biconditional statement
$(\neg A \vee B) \leftrightarrow P$,
we will call the auxiliary literal $P$ the consequent, and the
original literals $A$ and $B$ the antecedents.

Because the implication is bidirectional in biconditional statement,
it violates our acyclicity requirement in \S\ref{sec:workflow}.
However, since the auxiliary neuron state does not depend on any
other nodes,
we can still create an acyclic sub-graph by defining the new node to
be the distance function itself.


\paragraph{Constrained Auxiliary Layers}
With a biconditional statement $Z \leftrightarrow Y$, where $Y$ is
an auxiliary literal, we define a \emph{constrained auxiliary
  layer} as
\begin{equation}
  \label{eq:constrained-intermediate-layer}
  y = d\p{\bz}
\end{equation}
where $d$ is the distance function for the statement,
$\bz$ are upstream neurons associated with $Z$, $y$ is the
downstream neuron associated with $Y$. Note that, compared to
Eq.~\ref{eq:constrained-neural-layer}, we do not need activation
function since the distance, which is in $[0,1]$, can
be interpreted as producing normalized scores.

Note that this construction only applies to auxiliary predicates in biconditional statements.
The advantage of this layer definition is that we can use the same
distance functions as before (\ie,
Table~\ref{tab:tab1}). Furthermore, the same design considerations in
\S\ref{sec:workflow} still apply here, including how to negate the left and right hand sides.




\paragraph{Constructing augmented networks}
To complete the modeling framework, we summarize the workflow needed
to construct an augmented neural network given a conditional
statement and a computation graph:
\begin{inparaenum}[(1)]
\item Convert the antecedent into a conjunctive or a disjunctive
  normal form if necessary.
\item Convert the conjunctive/disjunctive antecedent into distance functions using
  Table~\ref{tab:tab1} (with appropriate corrections for negations).
\item Use the distance functions to construct constrained layers and/or auxiliary layers
  to augment the computation graph by replacing the original layer with constrained one.
\item Finally, use the augmented network for end-to-end training and
  inference.
\end{inparaenum}
We will see complete examples in \S\ref{sec:experiments}.

\subsection{Discussion}
\label{sec:combine}

Not only does our design not add any more trainable
parameters to the existing network, it also admits efficient
implementation with modern neural network libraries.


When posing multiple constraints on the same downstream neuron,
there could be combinatorial conflicts.  In this case, our framework
relies on the base network to handle the consistency
issue. In practice, we
found that summing the constrained pre-activation scores for
a neuron is a good heuristic (as we will see in \S\ref{sec:tc}).

For a conjunctive consequent, we can decompose it into multiple individual constraints.
That is equivalent to constraining downstream nodes independently.
Handling more complex consequents is a direction
of future research.



\section{Experiments}
\label{sec:experiments}

In this section, we will answer the research questions raised in
\S\ref{sec:intro} by focusing on the effectiveness of our
augmentation framework.  Specifically, we will explore three types
of constraints by augmenting: 1) intermediate decisions
(i.e. attentions); 2) output decisions constrained by intermediate
states; 3) output decisions constrained using label
dependencies. 



To this end, we instantiate our framework on three tasks: machine
comprehension, natural language inference, and text chunking.
Across all experiments, our goal is to study the modeling
flexibility of our framework and its ability to improve performance,
especially with decreasing amounts of training data. 

To study low data regimes, our augmented networks are trained using
varying amounts of training data to see how performances vary from
baselines. For detailed model setup, please refer to the appendices.

\subsection{Machine Comprehension}
\label{sec:mc}
Attention is a widely used intermediate state in several recent neural models.
To explore the augmentation over such neurons, we focus on attention-based machine comprehension models on
SQuAD (v1.1) dataset~\cite{rajpurkar2016squad}.
We seek to use word relatedness from external resources (\ie, ConceptNet) to guide alignments, and thus to improve model performance.



\paragraph{Model}
We base our framework on two models: BiDAF~\cite{seo2016bidirectional} and its ELMo-augmented variant~\cite{peters2018deep}.
Here, we provide an abstraction of the two models which our framework will operate on:
\begin{align}
  \bp, \bq &= \text{encoder}(p), \text{encoder}(q)\\
  \overleftarrow{\ba}, \overrightarrow{\ba} &= \sigma(\text{layers}(\bp, \bq)) \label{eq:bidafatt}\\
  \by, \bz &= \sigma(\text{layers}(\bp, \bq, \overleftarrow{\ba}, \overrightarrow{\ba}))
\end{align}
where $p$ and $q$ are the paragraph and query respectively, $\sigma$
refers to the softmax activation, $\overleftarrow{\ba}$ and
$\overrightarrow{\ba}$ are the bidirectional attentions from $q$ to
$p$ and vice versa, $\by$ and $\bz$ are the probabilities of answer
boundaries. All other aspects are abstracted as $encoder$ and
$layers$.

\paragraph{Augmentation}
%
By construction of the attention neurons, we expect that related
words should be aligned.
In a knowledge-driven approach, we can use ConceptNet to guide the
attention values in the model in Eq.~\ref{eq:bidafatt}.

We consider two rules to illustrate the flexibility of
our framework. Both statements are in first-order logic
that are dynamically grounded to the computation graph for a
particular paragraph and query. First, we define the following predicates:
\begin{tabular}{rp{0.8\linewidth}}
  $K_{i,j}$                        &word $p_i$ is related to word $q_j$ in ConceptNet via
                                     edges \{\emph{Synonym}, \emph{DistinctFrom},
                                     \emph{IsA}, \emph{Related}\}.           \\
  $\overleftarrow{A}_{i,j}$        &unconstrained model decision that
                                     word $q_j$
                                     best matches to word $p_i$. \\
  $\overleftarrow{A}^\prime_{i,j}$ &constrained model decision for
                                     the above alignment.
\end{tabular}
%
Using these predicates, we will study the impact of the following
two rules, defined over a set $C$ of content words in $p$ and $q$:
%
\begin{tabular}{rp{0.7\linewidth}}
  $R_1$: & $\forall i,j \in C,~ K_{i,j} \rightarrow\overleftarrow{A}^\prime_{i,j}$.  \\
  $R_2$: & $\forall i,j \in C,~ K_{i,j} \wedge \overleftarrow{A}_{i,j} \rightarrow \overleftarrow{A}^\prime_{i,j}$.
\end{tabular}

\begin{figure}[h]
  \centering
  \includegraphics[width=\linewidth]{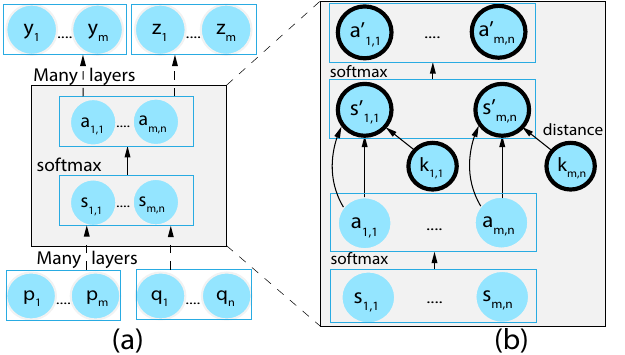}
  \caption{ (a) The computation graph of BiDAF where attention directions are obmitted. 
    (b) The augmented graph on attention layer using $R_2$.
    Bold circles are extra neurons introduced.
    Constrained attentions and scores are $\ba^\prime$ and
    $\bs^\prime$ respectively.
    In the augmented model, graph (b) replaces the shaded part in (a).
    }
  \label{fig:mcflow}
\end{figure}

The rule $R_1$ says that two words should be aligned if they
are related. Interestingly, compiling this statement using the
distance functions in Table~\ref{tab:tab1} is essentially the same
as adding word relatedness as a static feature.
The rule $R_2$
is more conservative as it also depends on the unconstrained model decision.
%
In both cases, since $K_{i,j}$ does not map to a node in the
network, we have to create a new node $k_{i,j}$ whose value is
determined using ConceptNet, as illustrated in Fig.~\ref{fig:mcflow}.


\paragraph{Can our framework use rules over named neurons to improve model performance?}
The answer is yes.
We experiment with rules $R_1$ and $R_2$ on incrementally larger training data.
Performances are reported in Table~\ref{tab:mc} with comparison with baselines.
We see that our framework can indeed use logic to inform model
learning and prediction without any extra trainable parameters needed.
The improvement is particularly strong with small training sets.
With more data, neural models are less reliant on external
information. As a result, the improvement with larger datasets
is smaller.

\begin{table}[]
  \centering
  \setlength{\tabcolsep}{1.5pt}
  \begin{tabular}{rccc|cc}
    \toprule
    {\%Train}  & { BiDAF} & { +$R_1$} & { +$R_2$} & { +ELMo} & { +ELMo,$R_1$} \\
    \midrule
    {10\%} & 57.5  &  \textbf{61.5}  & 60.7 & 71.8 & \textbf{73.0}\\
    {20\%} & 65.7  &  \textbf{67.2}  & 66.6 & 76.9 & \textbf{77.7}\\
    {40\%} & 70.6  &  \textbf{72.6}  & 71.9 & 80.3 & \textbf{80.9}\\
    {100\%} & 75.7 &  \textbf{77.4}  & 77.0 & 83.9 & \textbf{84.1}\\
    \bottomrule
  \end{tabular}
  \caption{Impact of constraints on BiDAF. Each
    score represents the average span $F_1$ on our test set (i.e. official dev set) among $3$ random runs.
    Constrained models and ELMo models are built on top of BiDAF.
    We set $\rho=2$ for both $R_1$ and $R_2$ across all percentages.}
  \label{tab:mc}
\end{table}

\paragraph{How does it compare to pretrained encoders?}
Pretrained encoders (e.g. ELMo and BERT~\cite{devlin2018bert})
improve neural models with improved representations,
while our framework augments the graph using first-order logic.
It is important to study the interplay of these two orthogonal directions.
We can see in Table~\ref{tab:mc}, our augmented model consistently outperforms baseline even with the presence of ELMo embeddings.

\paragraph{Does the conservative constraint $R_2$ help?}
We explored two options to incorporate word relatedness; one is a straightforward constraint (i.e. $R_1$), another is its conservative variant (i.e. $R_2$).
It is a design choice as to which to use.
Clearly in Table~\ref{tab:mc}, constraint $R_1$ consistently outperforms its conservative alternative $R_2$, even though $R_2$ is better than baseline.
In the next task, we will see an example where a conservative constraint performs better with large training data.

%
%


\subsection{Natural Language Inference}
\label{sec:nli}
Unlike in the machine comprehension task, here we explore logic rules that bridge attention neurons and output neurons.
We use the SNLI dataset~\cite{bowman2015large}, and base our framework on
a variant of the decomposable attention~\cite[DAtt,][]{parikh2016decomposable}
model where we replace its projection encoder with bidirectional LSTM (namely L-DAtt).




\paragraph{Model}
Again, we abstract the pipeline of L-DAtt model, only focusing on layers which our framework works on.
Given a premise $p$ and a hypothesis $h$, we summarize the model as:
\begin{align}
  \bp, \bh                                  & = \text{encoder}(p), \text{encoder}(h)   \\
  \overleftarrow{\ba}, \overrightarrow{\ba} & = \sigma(\text{layers}(\bp, \bh)) \\
  \by                                       & = \sigma(\text{layers}(\bp,\bh,\overleftarrow{\ba},\overrightarrow{\ba}))
\end{align}
Here, $\sigma$ is the softmax activation, $\overleftarrow{\ba}$ and
$\overrightarrow{\ba}$ are bidirectional attentions, $\by$ are
probabilities for labels \emph{Entailment}, \emph{Contradiction}, and \emph{Neutral}.

\paragraph{Augmentation}
We will borrow the predicate notation defined in the machine comprehension task (\S\ref{sec:mc}),
and ground them on premise and hypothesis words, e.g. $K_{i,j}$ now denotes the relatedness between
premise word $p_i$ and hypothesis word $h_j$.
In addition, we define the predicate $Y_l$ to indicate that the label is $l$.
As in \S\ref{sec:mc}, we define two rules governing attention:

\begin{tabular}{rp{0.7\linewidth}}
  $N_1$: & $\forall i,j \in C,~ K_{i,j} \rightarrow {A}^\prime_{i,j}$. \\
  $N_2$: & $\forall i,j \in C,~ K_{i,j} \wedge {A}_{i,j} \rightarrow {A}^\prime_{i,j}$.
\end{tabular}
where $C$ is the set of content words. Note that the two constraints apply to both attention directions.

\begin{figure}[h]
  \centering
  \includegraphics[width=\linewidth]{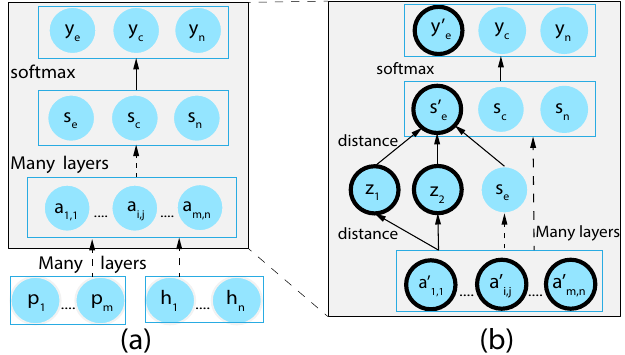}
  \caption{(a) The computation graph of the L-DAtt model (attention directions obmitted). 
    (b) The augmented graph on the $\emph{Entail}$ label using $N_{3}$.
    Bold circles are extra neurons introduced.
    Unconstrained pre-activation scores are $\bs$ while $\bs_e^\prime$ is
    the constrained score on \emph{Entail}.
    Intermediate neurons are $z_1$ and $z_2$.
    constrained attentions $\ba^\prime$ are constructed using $N_1$ or $N_2$.
    In our augmented model, the graph (b) replaces the shaded part in (a).
    }
  \label{fig:nliflow}
\end{figure}

Intuitively, if a hypothesis content word is not aligned, then the prediction should not be \emph{Entailment}.
To use this knowledge, we define the following rule:

\begin{tabular}{rp{0.7\linewidth}}
  $N_3$: & $Z_1 \wedge Z_2 \rightarrow \neg Y_{\text{Entail}}^\prime$, where\\
  & $\exists j \in C,~ \neg \p{\exists i \in C,~ \overleftarrow{A}^\prime_{i,j}} \leftrightarrow Z_1$,\\
  & $\exists j \in C,~ \neg \p{\exists i \in C,~ \overrightarrow{A}^\prime_{i,j}} \leftrightarrow Z_2$.
\end{tabular}
where $Z_1$ and $Z_2$ are auxiliary predicates tied to the $Y^\prime_{\text{Entail}}$ predicate.
The details of $N_3$ are illustrated in Fig.~\ref{fig:nliflow}.

\paragraph{How does our framework perform with large training data?}
The SNLI dataset is a large dataset with over half-million examples.
We train our models using incrementally larger percentages of data and report the average performance in Table~\ref{tab:nli}.
Similar to \S\ref{sec:mc}, we observe strong improvements from augmented models trained on small percentages ($\leq$10\%) of data.
The straightforward constraint $N_1$ performs strongly with
$\leq$2\% data while its conservative alternative $N_2$ works better
with a larger set.
However, with full dataset, our augmented models perform only on par with baseline even with lowered scaling factor $\rho$.
These observations suggest that if a large dataset is available, it
may be better to believe the data, but with smaller datasets,
constraints can provide useful inductive bias for the models.

\paragraph{Are noisy constraints helpful?}
It is not always easy to state a constraint that all examples satisfy.
Comparing $N_2$ and $N_3$, we see that $N_3$ performed even worse than baseline, which suggests it contains noise.
In fact, we found a significant amount of counter examples to $N_3$ during preliminary analysis.
Yet, even a noisy rule can improve model performance with $\leq$10\% data.
The same observation holds for $N_1$, which suggests conservative
constraints could be a way to deal with noise.
Finally, by comparing $N_2$ and $N_{2,3}$, we find that the good 
constraint $N_2$ can not just augment the network, but also amplify the noise in $N_3$ when they are combined.
This results in degrading performance in the $N_{2,3}$ column starting from 5\% of the data, much earlier than using $N_3$ alone.

\begin{table}[]
  \centering
  \setlength{\tabcolsep}{4pt}
  \begin{tabular}{rrrr|rr}
    \toprule
    \%Train    & L-DAtt & +$N_1$ & +$N_2$ & +$N_3$ & +$N_{2,3}$ \\
    \midrule
    1\%   & 61.2 & \textbf{64.9} & 63.9 & 62.5 &  64.3  \\
    2\%   & 66.5 & \textbf{70.5} & 69.8 & 67.9 &  70.2  \\
    5\%   & 73.4 & 76.2 & \textbf{76.6} &  74.0 &  76.4    \\
    10\%  & 78.9 & 80.1 & \textbf{80.4} &  79.3 &  80.3    \\
    100\% & \textbf{87.1} & 86.9  & \textbf{87.1} & 87.0 & 86.9     \\
    \bottomrule
  \end{tabular}
  \caption{Impact of constraints on L-DAtt network.
    Each score represents the average accuracy on SNLI test set among $3$ random runs.
    For both $N_1$ and $N_2$, we set $\rho=(8,8,8,8,4)$ for the five different percentages.
    For the noisy constraint $N_3$, $\rho=(2,2,1,1,1)$.
  }
  \label{tab:nli}
\end{table}


\subsection{Text Chunking}
\label{sec:tc}
Attention layers are a modeling choice that do not always exist in
all networks. To illustrate that our framework is not necessarily
grounded to attention, we turn to an application where we use
knowledge about the output space to constrain predictions.
We focus on the sequence labeling task of text chunking using the CoNLL2000 dataset~\cite{tjong2000introduction}.
In such sequence tagging tasks, global inference is widely used, e.g., BiLSTM-CRF~\cite{huang2015bidirectional}.
Our framework, on the other hand, aims to promote local decisions.
To explore the interplay of global model and local decision augmentation, we will combine
CRF with our framework.


\paragraph{Model} Our baseline is a BiLSTM tagger:
\begin{align}
  \bx &= \text{BiLSTM}(x) \\
  \by &= \sigma(\text{linear}(\bx))
\end{align}
where $x$ is the input sentence, $\sigma$ is softmax, $\by$ are the
output probabilities of BIO tags.

\paragraph{Augmentation}

We define the following predicates for input and output neurons:

\begin{tabular}{rp{0.75\linewidth}}
  $Y_{t,l}$ & The unconstrained decision that $t^{th}$ word has label $l$.\\
  $Y^\prime_{t,l}$ & The constrained decision that $t^{th}$ word has label $l$.\\
  $N_t$     & The $t^{th}$ word is a noun.
\end{tabular}

Then we can write rules for pairwise label dependency. For instance,
if word $t$ has B/I- tag for a certain label, word $t$+$1$ can not have an I- tag with a different label.

\begin{tabular}{rp{0.75\linewidth}}
  $C_1$: & $\forall t, ~Y_{t, \text{B-VP}}\rightarrow \neg Y^\prime_{t+1, \text{I-NP}}$  \\ 
  $C_2$: & $\forall t, ~Y_{t, \text{I-NP}} \rightarrow \neg Y^\prime_{t+1, \text{I-VP}}$ \\ 
  $C_3$: & $\forall t, ~Y_{t, \text{I-VP}} \rightarrow \neg Y^\prime_{t+1, \text{I-NP}}$ \\ 
  $C_4$: & $\forall t, ~Y_{t, \text{B-PP}} \rightarrow \neg Y^\prime_{t+1, \text{I-VP}}$  
\end{tabular}

Our second set of rules are also intuitive: A noun should not have non-NP label.
\begin{tabular}{rp{0.75\linewidth}}
  $C_5$: $\forall t, N_t \rightarrow  \bigwedge_{l\in \{\text{\small{B-VP,I-VP,B-PP,I-PP}}\}} \neg Y^\prime_{t, l}$ \\
\end{tabular}
While all above rules can be applied as hard constraints in the output space, our framework
provides a differentiable way to inform the model during training and prediction.

\paragraph{How does local augmentation compare with global inference?}
We report performances in Table~\ref{tab:tc}.
While a first-order Markov model (\eg, the BiLSTM-CRF) can learn pairwise constraints such as $C_{1:4}$,
we see that our framework can better inform the model.
Interestingly, the CRF model performed even worse than the baseline with $\leq$40\% data.
This suggests that global inference relies on more training examples to learn its scoring function.
In contrast, our constrained models performed strongly even with small training sets.
And by combining these two orthogonal methods, our locally augmented CRF performed the best with full data.

\begin{table}[]
  \centering
  \begin{tabular}{rcccc}
    \toprule
    {\small \%Train}   & {\small BiLSTM} & {\small +CRF} & {\small +$C_{1:5}$}  & {\small +CRF,$C_{1:5}$}     \\
	\midrule
	{5\%}   & 87.2   & 86.6 & \textbf{88.9} & 88.6 \\
	{10\%}  & 89.1   & 88.8 & \textbf{90.7} & 90.6 \\
	{20\%}  & 90.9   & 90.8 & \textbf{92.1} & \textbf{92.1} \\
	{40\%}  & 92.5   & 92.5 & 93.4 & \textbf{93.5} \\
	{100\%} & 94.1   & 94.4 & 94.8 & \textbf{95.0} \\
    \bottomrule
  \end{tabular}
  \caption{Impact of constraints on BiLSTM tagger.
    Each score represents the average accuracy on test set of $3$ random runs. 
    The columns of +CRF, +$C_{1:5}$, and +CRF,$C_{1:5}$ are on top of the BiLSTM baseline.
    For $C_{1:4}$, $\rho=4$ for all percentages. For $C_5$, $\rho=16$.}
  \label{tab:tc}
\end{table}



\section{Related Work and Discussion}
\label{sec:related}

\paragraph{Artificial Neural Networks and Logic}
Our work is related to neural-symbolic
learning~\cite[\eg][]{besold2017neural} which seeks to integrate
neural networks with symbolic knowledge. For
example,~\citet{cingillioglu2018deeplogic} proposed neural models
that multi-hop logical reasoning.

KBANN~\cite{towell1990refinement} constructs artificial neural
networks using connections expressed in propositional logic.  Along
these lines, \citet[CILP++]{francca2014fast} build neural networks
from a rule set for relation
extraction.  
Our distinction is that we use first-order
logic to {\em augment} a given architecture instead of designing a
new one. Also, our framework is related
to~\citet[PSL]{kimmig2012short} which uses a smooth extension of
standard Boolean logic.

~\citet{hu2016harnessing} introduced an imitation learning framework
where a specialized teacher-student network is used to distill rules
into network parameters.  This work could be seen as an instance of
knowledge distillation~\cite{hinton2015distilling}.  Instead of such
extensive changes to the learning procedure, our framework retains
the original network design and augments existing interpretable
layers.

\paragraph{Regularization with Logic}
Several recent lines of research seek to guide training neural
networks by integrating logical rules in the form of additional
terms in the loss functions~\cite[\eg,][]{rocktaschel2015injecting}
that essentially promote constraints among output
labels~\cite[\eg,][]{du2019consistent,mehta2018towards}, promote
agreement~\cite{hsu2018unified} or reduce inconsistencies across
predictions~\cite{minervini2018adversarially}.

Furthermore, \citet{pmlr-v80-xu18h} proposed a general design of
loss functions using symbolic knowledge about the outputs.
~\citet{fischer2019dl2} described a method for for deriving losses that are friendly to gradient-based learning algorithms. 
~\citet{wang2018deep} proposed a framework for integrating 
indirect supervision expressed via probabilistic logic into neural networks.


\paragraph{Learning with Structures}
Traditional structured prediction
models~\cite[\eg][]{smith2011linguistic} naturally admit constraints
of the kind described in this paper. Indeed, our approach for using
logic as a template-language is similar to Markov Logic
Networks~\cite{richardson2006markov}, where logical forms are
compiled into Markov networks.  Our formulation augments model
scores with constraint penalties is reminiscent of the Constrained
Conditional Model of \citet{chang2012structured}.

Recently, we have seen some work that allows backpropagating through
structures~\cite[\eg][and the references
within]{huang2015bidirectional,kim2017structured,yogatama2017learning,niculae2018sparsemap,peng2018backpropagating}.
Our framework differs from them in that structured inference is not mandantory here.
We believe that there is room to study the interplay of these two approaches.

Also related to our attention augmentation is using word relatedness as extra input feature to attention neurons~\cite[\eg][]{chen2018neural}.

\section{Conclusions}
\label{sec:conclusions}

In this paper, we presented a framework for introducing constraints
in the form of logical statements to neural networks.  We
demonstrated the process of converting first-order logic into
differentiable components of networks without extra
learnable parameters and extensive redesign.  Our experiments were
designed to explore the flexibility of our framework with different
constraints in diverse tasks.  As our experiments showed, our
framework allows neural models to benefit from external knowledge during learning and prediction, especially when training data is limited.


\section{Acknowledgements}
We thank members of the NLP group at the University of Utah for
their valuable insights and suggestions; and reviewers for pointers
to related works, corrections, and helpful comments. We also
acknowledge the support of NSF SaTC-1801446, and gifts from Google
and NVIDIA.

\bibliographystyle{style/acl_natbib}
\bibliography{cited}

\begin{thebibliography}{46}
\expandafter\ifx\csname natexlab\endcsname\relax\def\natexlab#1{#1}\fi

\bibitem[{Anthony(2003)}]{anthony2003boolean}
Martin Anthony. 2003.
\newblock Boolean functions and artificial neural networks.
\newblock \emph{Boolean Functions}.

\bibitem[{Bahdanau et~al.(2015)Bahdanau, Cho, and Bengio}]{bahdanau2014neural}
Dzmitry Bahdanau, Kyunghyun Cho, and Yoshua Bengio. 2015.
\newblock Neural machine translation by jointly learning to align and
  translate.
\newblock \emph{International Conference on Learning Representations}.

\bibitem[{Besold et~al.(2017)Besold, Garcez, Bader, Bowman, Domingos, Hitzler,
  K{\"u}hnberger, Lamb, Lowd, Lima et~al.}]{besold2017neural}
Tarek~R Besold, Artur~d'Avila Garcez, Sebastian Bader, Howard Bowman, Pedro
  Domingos, Pascal Hitzler, Kai-Uwe K{\"u}hnberger, Luis~C Lamb, Daniel Lowd,
  Priscila Machado~Vieira Lima, et~al. 2017.
\newblock Neural-symbolic learning and reasoning: A survey and interpretation.
\newblock \emph{arXiv preprint arXiv:1711.03902}.

\bibitem[{Bowman et~al.(2015)Bowman, Angeli, Potts, and
  Manning}]{bowman2015large}
Samuel~R Bowman, Gabor Angeli, Christopher Potts, and Christopher~D Manning.
  2015.
\newblock A large annotated corpus for learning natural language inference.
\newblock In \emph{Proceedings of the 2015 Conference on Empirical Methods in
  Natural Language Processing}.

\bibitem[{Chandra and Harel(1985)}]{chandra1985horn}
Ashok~K Chandra and David Harel. 1985.
\newblock Horn clause queries and generalizations.
\newblock \emph{The Journal of Logic Programming}, 2.

\bibitem[{Chang et~al.(2012)Chang, Ratinov, and Roth}]{chang2012structured}
Ming-Wei Chang, Lev Ratinov, and Dan Roth. 2012.
\newblock Structured learning with constrained conditional models.
\newblock \emph{Machine learning}, 88.

\bibitem[{Chen et~al.(2018)Chen, Zhu, Ling, Inkpen, and Wei}]{chen2018neural}
Qian Chen, Xiaodan Zhu, Zhen-Hua Ling, Diana Inkpen, and Si~Wei. 2018.
\newblock Neural natural language inference models enhanced with external
  knowledge.
\newblock In \emph{Proceedings of the 56th Annual Meeting of the Association
  for Computational Linguistics (Volume 1: Long Papers)}.

\bibitem[{Cingillioglu and Russo(2019)}]{cingillioglu2018deeplogic}
Nuri Cingillioglu and Alessandra Russo. 2019.
\newblock Deeplogic: End-to-end logical reasoning.
\newblock \emph{AAAI 2019 Spring Symposium on Combining Machine Learning with
  Knowledge Engineering}.

\bibitem[{Devlin et~al.(2018)Devlin, Chang, Lee, and
  Toutanova}]{devlin2018bert}
Jacob Devlin, Ming-Wei Chang, Kenton Lee, and Kristina Toutanova. 2018.
\newblock Bert: Pre-training of deep bidirectional transformers for language
  understanding.
\newblock \emph{Proceedings of the 2018 Conference of the North American
  Chapter of the Association for Computational Linguistics: Human Language
  Technologies}.

\bibitem[{Du et~al.(2019)Du, Dalvi, Tandon, Bosselut, tau Yih, Clark, and
  Cardie}]{du2019consistent}
Xinya Du, Bhavana Dalvi, Niket Tandon, Antoine Bosselut, Wen tau Yih, Peter
  Clark, and Claire Cardie. 2019.
\newblock Be consistent! improving procedural text comprehension using label
  consistency.
\newblock In \emph{Proceedings of the 2019 Conference of the North American
  Chapter of the Association for Computational Linguistics: Human Language
  Technologies}.

\bibitem[{Fischer et~al.(2019)Fischer, Balunovic, Drachsler-Cohen, Gehr, Zhang,
  and Vechev}]{fischer2019dl2}
Marc Fischer, Mislav Balunovic, Dana Drachsler-Cohen, Timon Gehr, Ce~Zhang, and
  Martin Vechev. 2019.
\newblock Dl2: Training and querying neural networks with logic.
\newblock In \emph{International Conference on Machine Learning}.

\bibitem[{Fran{\c{c}}a et~al.(2014)Fran{\c{c}}a, Zaverucha, and
  Garcez}]{francca2014fast}
Manoel~VM Fran{\c{c}}a, Gerson Zaverucha, and Artur S~d'Avila Garcez. 2014.
\newblock Fast relational learning using bottom clause propositionalization
  with artificial neural networks.
\newblock \emph{Machine learning}, 94.

\bibitem[{Hinton et~al.(2015)Hinton, Vinyals, and Dean}]{hinton2015distilling}
Geoffrey Hinton, Oriol Vinyals, and Jeff Dean. 2015.
\newblock Distilling the knowledge in a neural network.
\newblock In \emph{Neural Information Processing Systems}.

\bibitem[{Hsu et~al.(2018)Hsu, Lin, Lee, Min, Tang, and Sun}]{hsu2018unified}
Wan-Ting Hsu, Chieh-Kai Lin, Ming-Ying Lee, Kerui Min, Jing Tang, and Min Sun.
  2018.
\newblock A unified model for extractive and abstractive summarization using
  inconsistency loss.
\newblock In \emph{Proceedings of the 56th Annual Meeting of the Association
  for Computational Linguistics (Volume 1: Long Papers)}.

\bibitem[{Hu et~al.(2016)Hu, Ma, Liu, Hovy, and Xing}]{hu2016harnessing}
Zhiting Hu, Xuezhe Ma, Zhengzhong Liu, Eduard Hovy, and Eric Xing. 2016.
\newblock Harnessing deep neural networks with logic rules.
\newblock In \emph{Proceedings of the 54th Annual Meeting of the Association
  for Computational Linguistics (Volume 1: Long Papers)}.

\bibitem[{Huang et~al.(2015)Huang, Xu, and Yu}]{huang2015bidirectional}
Zhiheng Huang, Wei Xu, and Kai Yu. 2015.
\newblock Bidirectional lstm-crf models for sequence tagging.
\newblock \emph{arXiv preprint arXiv:1508.01991}.

\bibitem[{Kim et~al.(2017)Kim, Denton, Hoang, and Rush}]{kim2017structured}
Yoon Kim, Carl Denton, Luong Hoang, and Alexander~M Rush. 2017.
\newblock Structured attention networks.
\newblock In \emph{International Conference on Learning Representations}.

\bibitem[{Kimmig et~al.(2012)Kimmig, Bach, Broecheler, Huang, and
  Getoor}]{kimmig2012short}
Angelika Kimmig, Stephen Bach, Matthias Broecheler, Bert Huang, and Lise
  Getoor. 2012.
\newblock A short introduction to probabilistic soft logic.
\newblock In \emph{Proceedings of the NIPS Workshop on Probabilistic
  Programming: Foundations and Applications}.

\bibitem[{Klement et~al.(2013)Klement, Mesiar, and Pap}]{klement2013triangular}
Erich~Peter Klement, Radko Mesiar, and Endre Pap. 2013.
\newblock \emph{Triangular norms}.
\newblock Springer Science \& Business Media.

\bibitem[{Le et~al.(2012)Le, Ranzato, Monga, Devin, Chen, Corrado, Dean, and
  Ng}]{le2011building}
Quoc~V Le, Marc'Aurelio Ranzato, Rajat Monga, Matthieu Devin, Kai Chen, Greg~S
  Corrado, Jeff Dean, and Andrew~Y Ng. 2012.
\newblock Building high-level features using large scale unsupervised learning.
\newblock In \emph{International Conference on Machine Learning}.

\bibitem[{Liu and Singh(2004)}]{liu2004conceptnet}
Hugo Liu and Push Singh. 2004.
\newblock {ConceptNet -- A Practical Commonsense Reasoning Tool-Kit}.
\newblock \emph{BT technology journal}, 22.

\bibitem[{Luong et~al.(2015)Luong, Pham, and Manning}]{luong2015effective}
Thang Luong, Hieu Pham, and Christopher~D Manning. 2015.
\newblock Effective approaches to attention-based neural machine translation.
\newblock In \emph{Proceedings of the 2015 Conference on Empirical Methods in
  Natural Language Processing}.

\bibitem[{Maass et~al.(1994)Maass, Schnitger, and Sontag}]{maass1994comparison}
Wolfgang Maass, Georg Schnitger, and Eduardo~D Sontag. 1994.
\newblock A comparison of the computational power of sigmoid and boolean
  threshold circuits.
\newblock In \emph{Theoretical Advances in Neural Computation and Learning},
  pages 127--151. Springer.

\bibitem[{Mehta et~al.(2018)Mehta, Lee, and Carbonell}]{mehta2018towards}
Sanket~Vaibhav Mehta, Jay~Yoon Lee, and Jaime Carbonell. 2018.
\newblock Towards semi-supervised learning for deep semantic role labeling.
\newblock In \emph{Proceedings of the 2018 Conference on Empirical Methods in
  Natural Language Processing}.

\bibitem[{Minervini and Riedel(2018)}]{minervini2018adversarially}
Pasquale Minervini and Sebastian Riedel. 2018.
\newblock Adversarially regularising neural nli models to integrate logical
  background knowledge.
\newblock In \emph{Proceedings of the 22nd Conference on Computational Natural
  Language Learning}.

\bibitem[{Niculae et~al.(2018)Niculae, Martins, Blondel, and
  Cardie}]{niculae2018sparsemap}
Vlad Niculae, Andr{\'e}~FT Martins, Mathieu Blondel, and Claire Cardie. 2018.
\newblock {SparseMAP: Differentiable sparse structured inference}.
\newblock In \emph{International Conference on Machine Learning}.

\bibitem[{Pan and Srikumar(2016)}]{PanSr2016}
Xingyuan Pan and Vivek Srikumar. 2016.
\newblock Expressiveness of rectifier networks.
\newblock In \emph{International Conference on Machine Learning}.

\bibitem[{Parikh et~al.(2016)Parikh, T{\"a}ckstr{\"o}m, Das, and
  Uszkoreit}]{parikh2016decomposable}
Ankur Parikh, Oscar T{\"a}ckstr{\"o}m, Dipanjan Das, and Jakob Uszkoreit. 2016.
\newblock A decomposable attention model for natural language inference.
\newblock In \emph{Proceedings of the 2016 Conference on Empirical Methods in
  Natural Language Processing}.

\bibitem[{Paszke et~al.(2017)Paszke, Gross, Chintala, Chanan, Yang, DeVito,
  Lin, Desmaison, Antiga, and Lerer}]{paszke2017automatic}
Adam Paszke, Sam Gross, Soumith Chintala, Gregory Chanan, Edward Yang, Zachary
  DeVito, Zeming Lin, Alban Desmaison, Luca Antiga, and Adam Lerer. 2017.
\newblock Automatic differentiation in pytorch.
\newblock In \emph{NIPS 2017 Autodiff Workshop}.

\bibitem[{Peng et~al.(2018)Peng, Thomson, and Smith}]{peng2018backpropagating}
Hao Peng, Sam Thomson, and Noah~A Smith. 2018.
\newblock {Backpropagating through Structured Argmax using a SPIGOT}.
\newblock \emph{Proceedings of the 56th Annual Meeting of the Association for
  Computational Linguistics (Volume 1: Long Papers)}.

\bibitem[{Pennington et~al.(2014)Pennington, Socher, and
  Manning}]{pennington2014glove}
Jeffrey Pennington, Richard Socher, and Christopher Manning. 2014.
\newblock Glove: Global vectors for word representation.
\newblock In \emph{Proceedings of the 2014 Conference on Empirical Methods in
  Natural Language Processing}.

\bibitem[{Peters et~al.(2018)Peters, Neumann, Iyyer, Gardner, Clark, Lee, and
  Zettlemoyer}]{peters2018deep}
Matthew Peters, Mark Neumann, Mohit Iyyer, Matt Gardner, Christopher Clark,
  Kenton Lee, and Luke Zettlemoyer. 2018.
\newblock Deep contextualized word representations.
\newblock In \emph{Proceedings of the 2018 Conference of the North American
  Chapter of the Association for Computational Linguistics: Human Language
  Technologies}.

\bibitem[{Radford et~al.(2017)Radford, Jozefowicz, and
  Sutskever}]{radford2017learning}
Alec Radford, Rafal Jozefowicz, and Ilya Sutskever. 2017.
\newblock Learning to generate reviews and discovering sentiment.
\newblock \emph{arXiv preprint arXiv:1704.01444}.

\bibitem[{Rajpurkar et~al.(2016)Rajpurkar, Zhang, Lopyrev, and
  Liang}]{rajpurkar2016squad}
Pranav Rajpurkar, Jian Zhang, Konstantin Lopyrev, and Percy Liang. 2016.
\newblock Squad: 100,000+ questions for machine comprehension of text.
\newblock In \emph{Proceedings of the 2016 Conference on Empirical Methods in
  Natural Language Processing}.

\bibitem[{Richardson and Domingos(2006)}]{richardson2006markov}
Matthew Richardson and Pedro Domingos. 2006.
\newblock Markov logic networks.
\newblock \emph{Machine learning}, 62.

\bibitem[{Rockt{\"a}schel et~al.(2015)Rockt{\"a}schel, Singh, and
  Riedel}]{rocktaschel2015injecting}
Tim Rockt{\"a}schel, Sameer Singh, and Sebastian Riedel. 2015.
\newblock Injecting logical background knowledge into embeddings for relation
  extraction.
\newblock In \emph{Proceedings of the 2015 Conference of the North American
  Chapter of the Association for Computational Linguistics: Human Language
  Technologies}.

\bibitem[{Rush et~al.(2015)Rush, Chopra, and Weston}]{rush2015neural}
Alexander~M Rush, Sumit Chopra, and Jason Weston. 2015.
\newblock A neural attention model for abstractive sentence summarization.
\newblock In \emph{Proceedings of the 2015 Conference on Empirical Methods in
  Natural Language Processing}.

\bibitem[{Russell and Norvig(2016)}]{russell2016artificial}
Stuart~J Russell and Peter Norvig. 2016.
\newblock \emph{{Artificial Intelligence: A Modern Approach}}.
\newblock Pearson Education Limited.

\bibitem[{Seo et~al.(2017)Seo, Kembhavi, Farhadi, and
  Hajishirzi}]{seo2016bidirectional}
Minjoon Seo, Aniruddha Kembhavi, Ali Farhadi, and Hannaneh Hajishirzi. 2017.
\newblock Bidirectional attention flow for machine comprehension.
\newblock \emph{International Conference on Learning Representations}.

\bibitem[{Smith(2011)}]{smith2011linguistic}
Noah~A Smith. 2011.
\newblock Linguistic structure prediction.
\newblock \emph{Synthesis lectures on human language technologies}, 4.

\bibitem[{Srivastava et~al.(2014)Srivastava, Hinton, Krizhevsky, Sutskever, and
  Salakhutdinov}]{srivastava2014dropout}
Nitish Srivastava, Geoffrey Hinton, Alex Krizhevsky, Ilya Sutskever, and Ruslan
  Salakhutdinov. 2014.
\newblock Dropout: A simple way to prevent neural networks from overfitting.
\newblock \emph{The Journal of Machine Learning Research}, 15.

\bibitem[{Tjong Kim~Sang and Buchholz(2000)}]{tjong2000introduction}
Erik~F Tjong Kim~Sang and Sabine Buchholz. 2000.
\newblock {Introduction to the CoNLL-2000 shared task: Chunking}.
\newblock In \emph{Proceedings of the 2nd Workshop on Learning Language in
  Logic and the 4th Conference on Computational Natural Language Learning}.

\bibitem[{Towell et~al.(1990)Towell, Shavlik, and
  Noordewier}]{towell1990refinement}
Geoffrey~G Towell, Jude~W Shavlik, and Michiel~O Noordewier. 1990.
\newblock Refinement of approximate domain theories by knowledge-based neural
  networks.
\newblock In \emph{Proceedings of the Eighth National Conference on Artificial
  Intelligence}.

\bibitem[{Wang and Poon(2018)}]{wang2018deep}
Hai Wang and Hoifung Poon. 2018.
\newblock Deep probabilistic logic: A unifying framework for indirect
  supervision.
\newblock In \emph{Proceedings of the 2018 Conference on Empirical Methods in
  Natural Language Processing}.

\bibitem[{Xu et~al.(2018)Xu, Zhang, Friedman, Liang, and Van~den
  Broeck}]{pmlr-v80-xu18h}
Jingyi Xu, Zilu Zhang, Tal Friedman, Yitao Liang, and Guy Van~den Broeck. 2018.
\newblock A semantic loss function for deep learning with symbolic knowledge.
\newblock In \emph{International Conference on Machine Learning}.

\bibitem[{Yogatama et~al.(2017)Yogatama, Blunsom, Dyer, Grefenstette, and
  Ling}]{yogatama2017learning}
Dani Yogatama, Phil Blunsom, Chris Dyer, Edward Grefenstette, and Wang Ling.
  2017.
\newblock Learning to compose words into sentences with reinforcement learning.
\newblock In \emph{International Conference on Machine Learning}.

\end{thebibliography}


\begin{thebibliography}{4}
\expandafter\ifx\csname natexlab\endcsname\relax\def\natexlab#1{#1}\fi

\bibitem[{Paszke et~al.(2017)Paszke, Gross, Chintala, Chanan, Yang, DeVito,
  Lin, Desmaison, Antiga, and Lerer}]{paszke2017automatic}
Adam Paszke, Sam Gross, Soumith Chintala, Gregory Chanan, Edward Yang, Zachary
  DeVito, Zeming Lin, Alban Desmaison, Luca Antiga, and Adam Lerer. 2017.
\newblock Automatic differentiation in pytorch.

\bibitem[{Pennington et~al.(2014)Pennington, Socher, and
  Manning}]{pennington2014glove}
Jeffrey Pennington, Richard Socher, and Christopher Manning. 2014.
\newblock Glove: Global vectors for word representation.
\newblock In \emph{EMNLP}, pages 1532--1543.

\bibitem[{Peters et~al.(2018)Peters, Neumann, Iyyer, Gardner, Clark, Lee, and
  Zettlemoyer}]{peters2018deep}
Matthew Peters, Mark Neumann, Mohit Iyyer, Matt Gardner, Christopher Clark,
  Kenton Lee, and Luke Zettlemoyer. 2018.
\newblock Deep contextualized word representations.
\newblock In \emph{NAACL}, volume~1, pages 2227--2237.

\bibitem[{Srivastava et~al.(2014)Srivastava, Hinton, Krizhevsky, Sutskever, and
  Salakhutdinov}]{srivastava2014dropout}
Nitish Srivastava, Geoffrey Hinton, Alex Krizhevsky, Ilya Sutskever, and Ruslan
  Salakhutdinov. 2014.
\newblock Dropout: A simple way to prevent neural networks from overfitting.
\newblock \emph{The Journal of Machine Learning Research}, 15(1):1929--1958.

\end{thebibliography}

\appendix
\section{Appendices}
\label{sec:appendix}

Here, we explain our experiment setup for the three tasks: machine comprehension, natural language inference, and text chunking.
For each task, we describe the model setup, hyperparameters, and data splits.

For all three tasks, we used Adam~\cite{paszke2017automatic} for
training and use 300 dimensional GloVe~\cite{pennington2014glove}
vectors (trained on 840B tokens) as word embeddings.

\subsection{Machine Comprehension}
The SQuAD (v1.1) dataset consists of $87,599$ training instances and $10,570$ development examples.
Firstly, for a specific percentage of training data, we sample from the original training set.
Then we split the sampled set into 9/1 folds for training and development.
The original development set is reserved for testing only. This is because that the official test set is hidden,
and the number of models we need to evaluate is impractical for accessing official test set.

In our implementation of the BiDAF model, we use a learning rate $0.001$ to train the model for $20$ epochs.
Dropout~\cite{srivastava2014dropout} rate is $0.2$.
The hidden size of each direction of BiLSTM encoder is $100$.
For ELMo models, we train for $25$ epochs with learning rate $0.0002$. The rest hyperparameters are the same as in ~\cite{peters2018deep}.
Note that we did neither pre-tune nor post-tune ELMo embeddings.
The best model on the development split is selected for evaluation. No exponential moving average method is used.
The scaling factor $\rho$'s are manually grid-searched in \{$1,2,4,8,16$\} without extensively tuning.

\subsection{Natural Language Inference}
We use Stanford Natural Language Inference (SNLI) dataset which has $549,367$ training, $9,842$ development, and $9,824$ test examples.
For each of the percentages of training data, we sample the same proportion from the orginal development set for validation.
To have reliable model selection, we limit the minimal number of sampled development examples to be $1000$.
The original test set is only for reporting.

In our implimentation of the BiLSTM variant of the Decomposable Attention (DAtt) model,
  we adopt learning rate $0.0001$ for $100$ epochs of training. The dropout rate is $0.2$.
The best model on the development split is selected for evaluation.
The scaling factor $\rho$'s are manually grid-searched in \{$0.5,1,2,4,8,16$\} without extensively tuning.

\subsection{Text Chunking}
The CoNLL2000 dataset consists of $8,936$ examples for training and $2,012$ for testing.
From the original training set, both of our training and development examples are sampled and split (by 9/1 folds).
Performances are then reported on the original full test set.

In our implementation, we set hidden size to $100$ for each direction of BiLSTM encoder.
Before the final linear layer, we add a dropout layer with probability $0.5$
for regularization. Each model was trained for $100$ epochs with learning rate $0.0001$.
The best model on the development split is selected for evaluation.
The scaling factor $\rho$'s are manually grid-searched in \{$1,2,4,8,16,32,64$\} without extensively tuning.

\end{document}